\documentclass{article}

\usepackage{microtype}
\usepackage{graphicx}
\usepackage{subfigure}
\usepackage[utf8]{inputenc} % allow utf-8 input
\usepackage{booktabs} % for professional tables
\usepackage{amsmath}
\usepackage{amsfonts}       % blackboard math symbols
\usepackage{todonotes}
\usepackage[subtle]{savetrees}
\usepackage{hyperref}

\usepackage[accepted]{icml2021}

\usepackage[belowskip=-15pt,aboveskip=0pt]{caption}

\icmltitlerunning{Principled Curriculum Learning }

\begin{document}
\newtheorem{theorem}{Theorem}[section]
\newtheorem{corollary}{Corollary}[theorem]
\newtheorem{lemma}[theorem]{Lemma}

\twocolumn[
\icmltitle{Principled Curriculum Learning using \\
           Parameter Continuation Methods}




\begin{icmlauthorlist}
\icmlauthor{Harsh Nilesh Pathak}{wpi,eg}
\icmlauthor{Randy Paffenroth}{wpiall}
\end{icmlauthorlist}
\icmlaffiliation{wpi}{Department of Data Science, Worcester Polytechnic Institute, USA}
\icmlaffiliation{eg}{Expedia Group, USA}
\icmlaffiliation{wpiall}{Department of Mathematics, Computer and Data Science, Worcester Polytechnic Institute, USA}
\icmlcorrespondingauthor{Harsh Nilesh Pathak}{hnpathak@wpi.edu, hpathak@expediagroup.com}
\icmlcorrespondingauthor{Randy Paffenroth}{rcpaffenroth@wpi.edu}

\icmlkeywords{Machine Learning, ICML}

\vskip 0.3in
]



\printAffiliationsAndNotice{\icmlEqualContribution} % otherwise use the standard text.

\begin{abstract}
In this work, we propose a parameter continuation method for the optimization of neural networks. There is a close connection between parameter continuation, homotopies, and curriculum learning. The methods we propose here are theoretically justified and practically effective for several problems in deep neural networks. In particular, we demonstrate better generalization performance than state-of-the-art optimization techniques such as ADAM for supervised and unsupervised learning tasks.

% In this work, we propose to study deep neural network optimization using numerical continuation methods, which are popular in the study of nonlinear dynamical systems. These methods have had many applications in physics and mathematical analysis for the study of the bifurcation structure of dynamical systems.  Herein, we study two well-known continuation methods, namely Natural and Pseudo-arclength continuation, and compare them with standard gradient descent based approaches. There is a close connection between parameter continuation, homotopies, and curriculum learning, and the methods we propose here provide a theoretically justified and practically effective for several problems in deep neural networks.  In particular, we demonstrate better generalization performance than state-of-the-art optimization techniques such as ADAM for supervised and unsupervised learning tasks.  
% In this paper, we propose a principled curriculum approach we provide this flexibility to the neural networks training process (via both data and model curriculum) and analyze the results in the experiment section. 

\end{abstract}

\section{Introduction}
Deep learning applications have seen remarkable progress in recent years \cite{deep_nature, Goodfellow-et-al-2016, 8622477}. However, the performance of neural networks is highly dependent on hyper-parameter choices such as loss function, network architecture design, activation function, training strategy, optimizer, initialization, and many other considerations. Unfortunately, many of these choices can lead to highly non-convex optimization problems that then need to be solved for the training of the deep neural network.  Another domain in which highly non-convex problems arise is dynamical systems. In fact, the word ``chaotic'' \cite{kathleen1997chaos} has become synonymous with the properties of some such systems.  \emph{Accordingly, herein we draw inspiration from the study of dynamical systems and propose to analyze deep neural networks from the perspective of homotopy methods and parameter continuation algorithms.}

For our proposed training methods, we transform standard deep neural networks using homotopies \cite{pathak2025soloconnectionparameterefficient, pathak2018npacs, pathak2024neural_phd, pathak2018parameter}. Such homotopies allow one to decompose the complex optimization problem into a sequence of simpler problems, each of which is provided with a good initial guess based upon the solution of the previous problem. Accordingly, in this work, we show how one can analyze the evolution of extrema based on the numerical continuation of some homotopy parameter for neural networks. These concepts are not new \cite{cont_method_Allgower} and have been used in other fields of mathematics such as discrete and continuous dynamical systems.  However, these techniques have not been widely used for analyzing deep neural networks even though they provide many advantages.

\subsection{Standard Training for Neural Networks}
\begin{equation}\label{eq:nn_loss}
    \theta^* = \underset{\theta}{argmin} \hspace{0.2cm}  
    \frac{1}{N} \sum_{i=1}^{N} L(y_i ,\hat{y_i})
\end{equation}

Given a task, dataset and a network architecture the standard techniques for training the neural network is to apply optimization techniques to a problem similar to the one given by equation (\ref{eq:nn_loss}), where  $\hat{y} = f(x; \theta)$ is the output of the neural network.  Classically, a variant of a minibatch gradient descent optimizer, perhaps with momentum term \cite{adagrad, hinton2012rmsprop, adam}, is iteratively applied to find the optimal network parameters. Unfortunately, a deep neural network's cost surface usually consists of many critical points \cite{Goodfellow-et-al-2016} such as local minima, saddle points and degenerate minima and saddle region. Thus, getting to the quality minimum with very low generalization performance is an active area of research.

% , as shown in equation (\ref{eq:gd_update}).  In this equation, $B$ denotes a batch from the dataset. 
% \begin{equation}\label{eq:gd_update}
%     \theta^{k+1} = \theta^{k} - \alpha  \frac{1}{B_k} \sum_{i \in B_k} \nabla L_i(\theta)
% \end{equation}

\subsection{Continuation Methods for Neural Networks}

Parameter continuation methods \cite{cont_method_Allgower, soviany2022curriculum, pathak2018parameter} take a different approach than the standard training. As introduced, continuation methods utilize homotopies to decompose the original problem to a continuum of tasks to work with a family of minima and thus, starts by finding a minimum (or critical) point for the simpler optimization problem.  Then, \emph{the optimization problem is gradually changed from the easy problem to the challenging problem of interest}.  The critical point is adjusted as the optimization problem is changed, leading to finding a critical point of the challenging problem.  

In particular, given a challenging minimization problem $\underset{\theta \in \mathbb{R}^m}{min} L(\theta)$, one can embed this problem into a larger class of problems using a \emph{homotopy} such as :
\begin{equation}\label{eq:min_p}
    \underset{\theta \in \mathbb{R}^m, \lambda \in [0,1]}{min} \lambda L(\theta) + (1-\lambda) M(\theta) 
\end{equation}
 
\noindent where $M(\theta)$ is some problem where a good initialization $\theta^0$ is known, which is in the basin of attraction of some critical point.
Given $\theta^0$ and setting $\lambda=0$ the above optimization problem can be solved using any first or second order gradient methods to converge at the critical point. The parameter $\lambda$ can then be ``continued in'' by increasing in small steps until $\lambda=1$ is reached and a critical point is found of the problem of interest $L$.

Of course, many questions present themselves.  Under what circumstances can we guarantee that we will eventually find a solution where $\lambda=1$?  What is an appropriate ``small'' step size?  What if $L$ has many critical points and $M$ has only one?  Such questions are precisely those that arise and are addressed by continuation method theory. Such homotopy embedding and continuation methods have long served as useful tools in modern mathematics \cite{cont_method_Allgower, klein1883neue,poincar, leray1934topologie}. The use of deformations to solve nonlinear systems of equations may be traced back at least to \cite{lahaye1934methode}. 

% \textbf{Contributions } \todo{Write little better}\\
% We empirically show for a supervised and an unsupervised task our Principled Curriculum Learning strategy provides better generalization results. We present a continuation framework to unify  the model and data curriculum for the different kinds of neural network.

% In this paper, we demonstrate the use of local continuation methods \cite{Keller77, cont_method_Allgower} to find the global solution of the systems such as equation \ref{eq:min_p} by parameterize this optimization using, a continuation parameter $\lambda$. This is the initial setup for continuation methods. Now suppose we are given a solution $(\theta^0, \lambda^0)$ of \ref{eq:cont_sys}. The idea of local continuation is to find a solution at $\lambda^0 + \delta \lambda$ for a small perturbation $\delta \lambda$. Then perhaps, we can proceed step by step, to get a global solution. 
% \begin{equation}
%     P(\lambda) := \underset{\theta \in R^m}{min} L(\theta, \lambda) 
% \end{equation}

\section{Curriculum and Continuation Methods}

We discussed two approaches to solving a non-convex optimization problem; First, a direct method where data is fed randomly and initialization of parameters is also random. Second, a continuation-based approach where we start with a simpler (possibly convex) problem which is gradually transformed to the highly non-convex problem. In this section, we want to shed some light on another popular approach which is originally inspired by continuation methods i.e. curriculum learning \cite{Bengio_curriculumlearning}. In general, curriculum learning suggests feeding data in a meaningful order; similar to humans who learn the tasks with increasing difficulty. Many researchers observed better generalization performance after introducing curriculum strategies to existing SOTA neural architectures \cite{soviany2021curriculum, power_cl_guy, progressive_gans, CLbyTL, poet_jeff}. The authors \cite{soviany2021curriculum, pathak2018parameter} broadly classify curriculum strategies as, (1) by using meaningful order of samples \emph{(data curriculum)}, and (2) by altering some carefully chosen model configuration \emph{(model curriculum)}. For a detailed study on recent curriculum strategies we recommend this paper \cite{soviany2021curriculum}.

Despite the better performance of curriculum learning, it has not been widely accepted by the Deep Learning community \cite{soviany2021curriculum}. Even in NLP, Active learning is more popular \cite{chandrasekaran2020deep}. One of the possible reasons that the curriculum needs to integrate well the in-hand optimization task is that the difficulty of devising such strategies may be domain-dependent and may also require careful human intervention. However, instead of one's intuition, we study curriculum strategies through the lens of Implicit Function Theorem (IFT) \cite{cont_method_Allgower}. In this paper, we attempt to close the gap between curriculum learning and continuation methods. In particular, if we define a single parameter $\lambda$ to employ data or model curriculum, then we discuss: 

\emph{Question: What is the best parametrization ($\lambda$) to find a family of minima for complex problems like Neural Networks?} In the recent literature, researchers have chosen several directions for applying curriculum learning in neural network training. Noisy activation \cite{Noisy_GulcehreMDB16} and Homotopy activation \cite{pathak2018npacs, pathak2023sequentia12d, pathak2024neural_phd, nilesh2020non} have been used to continue from linear to non-linear networks gradually. Anneal smoothing in convolution layers \cite{smooth2020curriculum}  and modify keep-probability \cite{dropout2017curriculum} in neural networks have been used to condition training at earlier epochs. In addition to these model variations, researchers have observed empirical performance gains when SOTA networks are trained with data curriculum rather than usual random shuffling \cite{soviany2021curriculum}. In most cases, these special parameters are updated manually or adaptively based on some performance measure. Numerical continuation theory provides powerful tools such as re-parameterization of control parameter ($\lambda$) and the IFT according to which we conjecture the following - To solve the continuum of tasks, more than the selected class of ($\lambda$); the formalization of how we parameterize the progress along the continuum of tasks is vital. \newline \emph{Answer (Informal): For non-convex problems, it turns out there is no single ($\lambda$) that you can smoothly parameterize all the families of minima for a Neural Network. Accordingly, the principled way would be to re-parameterize your problem using an intrinsic property of family of minima which, in our case, is the arclength parameter ($s$). }

 In the 1970s, the exact re-parameterization that we require for Neural Networks, namely Pseudo-arclength Continuation (PARC) \cite{KELLER197873}, was discovered.  It was the first robust technique to parameterize all the families of minima for complex problems. This method is being used to date in many mathematical software packages such as AUTO \cite{2007auto}, MATCONT \cite{MATCONT}, LOCA \cite{trilinos}, PyDSTool \cite{clewley2007pydstool}, etc. We explain more details on PARC in the next section.

% m   solve  and  this one of the pioneers of 
% From numerical continuation theory, there are no special characteristics of $\lambda$.  However, we want to observe the dynamics of such methods purely from the lens of IFT, which is a primary tool to study continuation methods and bifurcation analysis \cite{Keller77, cont_method_Allgower}. \todo{read carefully to here}

%\textbf{For example, GAN this, images that BLAH BLAH.}

%\textbf{However, parameter continuation methods provide a framework for doing curriculum learning that is less ad hoc.  Consider...}

\section{Continuation on solution path}
% {\color{blue}Until now we have understood to devise the continuum of tasks for neural networks. Now we discuss the details about the structure and dimension of this continuum. Also, we will illustrate the mapping of the set of solutions for the set of optimization tasks.}

We define the homotopy between an easier and a complex optimization problem by adding a single parameter $\lambda$ such that our new optimization problem is $\Tilde{L}(\theta,\lambda) = \lambda L(\theta) + (1-\lambda) M(\theta) $. For such a system we will get a set of solutions represented by the implicit relation $\theta(\lambda)$ \cite{cont_method_Allgower}. To solve this minimization problem $\Tilde{L}(\theta,\lambda)$, one way is to find the solutions or roots to the critical point equation. 
\begin{equation}
    H(\theta, \lambda) = \nabla_{\theta} \Tilde{L}(\theta, \lambda) = \textbf{0}
\end{equation}
where, $\quad H: \mathbb{R}^{n} \times \mathbb{R} \rightarrow \mathbb{R}^{n}, \quad \lambda \in \mathbb{R}$ such that $\Tilde{L}(\theta, 0) = M(\theta)$ a trivial problem and $\Tilde{L}(\theta, 1) = L(\theta)$ a non-trivial problem, as shown in equation \ref{eq:min_p}. By IFT, if a regular solution of $H$ is known at $(\theta_0, 0)$ then a smooth \textbf{\emph{solution path}} or curve exists in that neighbourhood and passes through $(\theta_0, 0)$. An example of a solution path is shown in Figure \ref{fig:parc}.

% There are two well-studied path following methods in continuation literature \cite{cont_method_Allgower} (1) Natural Parameter Continuation (2) Pseudo-arclength Continuation. These methods follow a predictor-corrector scheme and are employed in software packages such as Auto-07p \cite{2007Auto} and Auto-2000 \cite{2001Auto} available for Ordinary Differential Equations (ODEs). 

% Here, we want to start from a easier version of the problem i.e. our initialization is already in the basin of attraction of a some critical point.  Now instead of getting one optimal minimum as $\theta^*$, in the case of parameterized problem,  

\begin{theorem}
\textbf{Implicit Function Theorem (IFT) \cite{cont_method_Allgower}}

Let $H: \mathbb{R}^{n} \times \mathbb{R}^{p} \rightarrow \mathbb{R}^{n}$ be a $C^{1}$ -function $H(\theta, \lambda) .$ Suppose, \\
    1.  $H(\bar{\theta}, \bar{\lambda})=0$ ; for $(\bar{\theta}, \bar{\lambda}) \in \mathbb{R}^{n} \times \mathbb{R}^{p}$ \\
    2. $\nabla_{\theta} H(\bar{\theta}, \bar{\lambda}) \quad$ is nonsingular. (a.k.a regularity condition)

Then there exists a neighborhood $B_{\varepsilon}(\bar{\lambda}), \varepsilon>0$, of $\bar{\lambda}$ and a $C^{1}$ -function $\theta: B_{\varepsilon}(\bar{\lambda}) \rightarrow$
$\mathbb{R}^{n}$ satisfying $\theta(\bar{\lambda})=\bar{\theta}$ such that near $(\bar{\theta}, \bar{\lambda})$ the solution set $S(H):=\{(\theta, \lambda) \mid$ $H(\theta, \lambda)=0\}$ is described by parameterized form $\left\{(\theta(\lambda) \in \mathbb{R}^{n} \times \mathbb{R}^{p} \mid \lambda \in B_{\varepsilon}(\bar{\lambda})\right\}, \quad$ i.e., $\quad H(\theta, \lambda))=0$ for $\lambda \in B_{\varepsilon}(\bar{\lambda})$.
So, locally near $(\bar{\theta}, \bar{\lambda})$, the set $S(H)$ is a p-dimensional $C^{1}$ -manifold (a.k.a solution path).
\end{theorem}
% Moreover, the gradient $\nabla \theta(\lambda)$ is given by
% $$
% \nabla \theta(\lambda)=-\left[\nabla_{x} H(\theta, \lambda)\right]^{-1} \nabla_{\lambda} H(\theta, \lambda) \quad \text { for } \lambda \in B_{\varepsilon}(\bar{\lambda}) .
% $$

\emph{Our prime contribution is to rethink neural network training as tracing a solution path from an easy optimization to a highly non-convex optimization problem,  rather than direct solvers such as ADAM with random initialization.} Closely tracing such locally existent solution paths can be interpreted as always having a good initialization for each of the harder problems on the path of minima. Say, we know $\theta_0$ is very close for the solution at $\lambda_1$. Then we can easily solve to get $\theta_1$, since $\theta_0$ is in the basin of attraction for the problem defined by $\lambda_1$. Similarly we use $\theta_1$ to find the solution at $\lambda_2$. In other words, efficient tracing methods may remain in the basin of attraction, if they follow the solution path closely. However, tracing is a difficult task in high dimensional dynamical systems. The IFT teaches us that the solution path is smooth and unique locally. However, to the best of our knowledge there are no such claims on the global structure of the solution path. Especially when the regularity condition fails  or ($\nabla_{\theta} H(\bar{\theta}, \bar{\lambda}) $) is singular, then the solution path may show some singularity (bifurcations \footnote{Gradient descent iterations can be seen as iterative dynamical systems, where bifurcations are sudden behavioural change in parameters space at particular points.} \cite{cont_method_Allgower, nilesh2020non, pathak2024neural_phd} or non-smoothness). This introduces challenges to trace the solution path closely as you can no longer define your solution path with the natural parameter $\lambda$. As shown in Figure \ref{fig:parc} when the solution path folds onto itself. As a result, we fail to remain in the basin of attraction and might not converge at all for the respective task in  the continuum. 
For example, one dimensional Logistic Map \cite{1976Nature, kathleen1997chaos} is well known dynamical system with several limit points.  In order to mitigate this problem, the science behind the arclength parameter is helpful to perform robust continuation. Originally, tracing is performed using newton's method which is efficient for low dimensional problems, as it involves computing of the Hessian. In the case of deep learning, we usually train millions of parameters and computing Hessian can be very expensive, hence we develop these paths following methods combining gradient properties, matrix-free and algebraic methods. To get a overview on path tracing methods we suggest this book \cite{cont_method_Allgower}.

\section{Method: Pseudo-arclength continuation (PARC) for high dimensional problems}
%\subsection{Natural Parameter Continuation (NPC)}
%\todo{this should be the start of Section 3}

In order to include parameter $\lambda$ in the neural network optimization, we propose two homotopies (1) Activation Homotopy and (2) Brightness Homotopy. This is an elementwise operation on a input matrix. Example: $h(z) = (1-\lambda) \cdot z + \lambda \cdot sigmoid(z)$, we refer these as h-sigmoid in experimental results. Through this formulation we achieve the decomposition of neural network optimization to several tasks, for which we will now construct a path-following strategy. 

The simplest way to approximately follow a solution curve $\theta(\lambda)$ of $H(\theta, \lambda)=0$, on an interval $\lambda \in[a, b]$ is to discretize $[a, b]$ by
\begin{equation}
\lambda_{\ell}=a+\ell \frac{b-a}{N}, \quad \ell=0, \ldots, N
\end{equation}
for some $N \in \mathbb{N}$. 

 The tracing is carried out in two steps: predictor and corrector. Predictor computes the next difficulty level $\lambda_1 = \lambda_0+\Delta \lambda$, and corrector using the solution at $\theta(\lambda_0) = \theta_0$ solves for the new problem at $\lambda_1$; using any first-order gradient methods. The solution path following strategy could iteratively perform this predictor-corrector scheme to find a solution at $\lambda_n=1$ (non-trivial problem). This strategy is known as Natural Parameter Continuation (NPC) method, and explained in greater details in literature \cite{Keller77, cont_method_Allgower}. Recently, \cite{pathak2018npacs} adopted and modified NPC to work with neural networks (Autoencoders) and observed better convergence performance than most direct solvers.

%  for example, Pitchfork Bifurcation \cite{cont_method_Allgower}
% \begin{equation}\label{eq:fork}
% P(\lambda) \quad \min _{\theta \in \mathbb{R}} f(\theta, \lambda)= \frac{\theta^4}{4}+\lambda \frac{\theta^2}{2}, \quad \text {for} \quad \lambda \in \mathbb{R},
% \end{equation}

% Herein the critical point equation $H(\theta, \lambda) = \theta^3 + \lambda \theta =0$, we find at $\bar{\lambda} = 0$ the curve is smooth but in this special case we have one solution branch near $\lambda=0^{+}$ and three fixed/critical points near $\lambda=0^{-}$. In this case it is simple as even if we are able to follow just one of the evolving solution path the objective is minimized. However we are not sure how such scenarios occur in neural network training, we don't know how many times these singularities  will arise and finally if there are multiple solution paths which branch will provide best generalization. This raises an important question, how we evolve path tracking method to follow multiple branches; so that you can evaluate each branch to measure generalization performance. 

% \textbf{Example 1: Monotonic solution path}
% \begin{equation}\label{eq:simple_path}
% P(\lambda) \quad \min _{x \in \mathbb{R}} f(x, t)= t + x^{2}, \quad \text { for } \quad t \in \mathbb{R},
% \end{equation}  Then the solution path is a straight line $x=0$ i.e $S(t) = 0$ for $t \in R$

\begin{figure}
    \centering
    \includegraphics[scale=0.35]{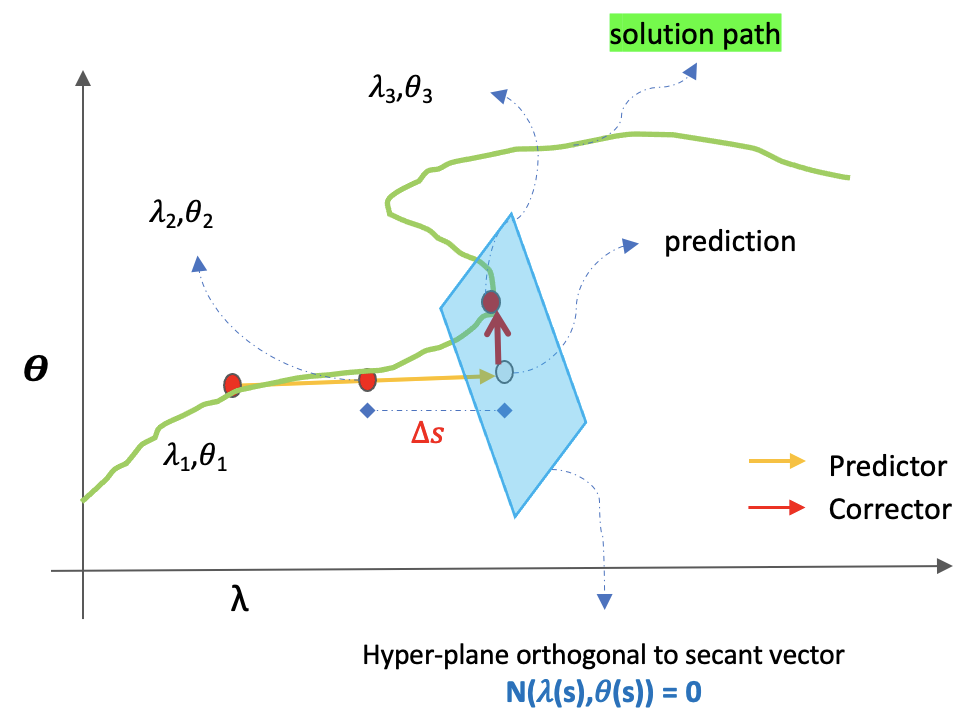}
    \caption{Pseudo-arclength Continuation}
    \label{fig:parc}
\end{figure}

However, NPC is not suitable when solution paths are not monotonic to predict. The solution path may consist of singularities such as folds (points which cannot be parameterized by $\lambda$) and bifurcations. To mitigate this issue we propose a more principled predictor-corrector framework to provide a robust tracking around singularities in solution paths. Pseudo-arclength Continuation (PARC) for Neural Networks is the main contribution of this paper. Originally, PARC use second and third order derivatives \cite{cont_method_Allgower} which is a major computational concern in high dimensions. Hence, we developed a first-order version of PARC that also uses some matrix-free methods such as secant to efficiently track the solution path and present a simplified algorithm for the same. In our case, the solution path is not tracked using $\lambda$ as the main continuation parameter. Instead, an intrinsic property of the solution path, the arclength $s$ (the distance you travel on the solution path) is used, such that, $(\theta(s), \lambda(s))$. This allows one to construct a robust tracing method. As illustrated in Figure \ref{fig:parc}, we first use a secant predictor to progress arclength by $\Delta s$ and then true network parameters are searched at fixed arclength. Specifically, the solution to harder problems is not searched at fixed $\lambda$ but at fixed $s$, while $\lambda$ is simultaneously adapted using corrector. Corrector uses solver methods such as ADAM or Newton on the regular loss with additional orthogonal constraint. In particular, gradient descent updates are penalized for moving out from the hyperplane orthogonal to the secant. This ensures to closely trace the solution paths with folds and hence initialization may always remain in the basin of attraction for all family of minima. Our version of PARC is also able to track multiple solution paths in case of bifurcations, but we limit our scope to the idea of continuation, solution paths and arclength parameterization for this paper.

%This method was usually used in a low dimensional application in the Chemical, Biology, and mechanical fields.  

% In your example_paper.tex preamble:
% REMOVE/COMMENT OUT: \usepackage{algorithm2e}
% MAKE SURE THE TEMPLATE LOADS: \RequirePackage{algorithm} and \RequirePackage{algorithmic}

% In your example_paper.tex preamble:
% REMOVE/COMMENT OUT: \usepackage{algorithm2e}
% MAKE SURE THE TEMPLATE LOADS: \RequirePackage{algorithm} and \RequirePackage{algorithmic}
% ALSO ADD (if not already present and if you want \PROCEDURE):

% Your algorithm in example_paper.tex:
% In your example_paper.tex preamble:
% REMOVE/COMMENT OUT: \usepackage{algorithm2e}
% REMOVE/COMMENT OUT: \usepackage{algpseudocode}
% MAKE SURE THE TEMPLATE LOADS: \RequirePackage{algorithm} and \RequirePackage{algorithmic}

% In your example_paper.tex preamble:
% REMOVE/COMMENT OUT: \usepackage{algorithm2e}
% REMOVE/COMMENT OUT: \usepackage{algpseudocode}
% MAKE SURE THE TEMPLATE LOADS: \RequirePackage{algorithm} and \RequirePackage{algorithmic}

% Your algorithm in example_paper.tex:
\begin{algorithm}
\caption{Pseudo-arclength Continuation} % Use \caption for the title
\label{algo:parc} % \label after \caption

\begin{algorithmic}[1] % [1] for line numbers, omit if you don't want them
    
    \STATE global\_list = []

    \STATE Loss with orthogonality constraint:
    \STATE $\Tilde{L}(\theta, \lambda) = \frac{1}{N} \sum ||\hat{y}-y||_F^2 + \gamma (\Delta \theta \cdot \Dot{\theta} + \Delta \lambda \cdot \Dot{\lambda}) $

    % --- Predictor Function ---
    \STATE \COMMENT{Function predictor($\theta, \lambda$):} % Using \COMMENT for a clearer label
    \STATE \hspace{0.2in} $\theta = \theta + \frac{(\theta - \theta_{-1}) \Delta s}{||\Delta \theta||} $
    \STATE \hspace{0.2in} $\lambda = \lambda + \frac{(\lambda - \lambda_{-1}) \Delta s}{|\Delta \lambda|} $
    \STATE \hspace{0.2in} \COMMENT{Return ($\theta, \lambda$)} % Using \COMMENT for return statement

    % The empty \STATE here adds an extra blank line, you might want to remove it
    % if you prefer less vertical space between functions.
    \STATE % This line was in your example; remove if not desired.

    % --- Corrector Function ---
    \STATE \COMMENT{Function corrector($\theta, \lambda$):} % Using \COMMENT for a clearer label
    \STATE \hspace{0.2in} Say, init with $(\theta_p, \lambda_p)$
    \STATE \hspace{0.2in} on orthogonal-plane to the secant vector.
     \WHILE{convergence}
        % Note: The \WHILE block itself adds an indent.
        % Adding another \hspace{0.2in} here will indent it further.
        \STATE \hspace{0.2in} $\theta = \theta - \alpha \nabla \Tilde{L}(\theta, \lambda)$
    \ENDWHILE
    \STATE \hspace{0.2in} \COMMENT{Return ($\theta, \lambda$)} % Using \COMMENT for return statement

    % --- Main Execution Block ---
    \STATE 
    \STATE \COMMENT{Main Execution Block}
    \STATE $\lambda = \lambda_0$, $\theta = \theta_0$
    \WHILE{$\lambda <= 1$}
        \STATE $\theta$,$\lambda$ = predictor($\theta, \lambda$)
        \STATE $\theta$,$\lambda$ = corrector($\theta, \lambda$)
    \ENDWHILE
    \STATE Append ($\theta$,$\lambda$) to global\_list
\end{algorithmic}
\end{algorithm}

\section{Experiments}
In this section, we present results on neural networks. We performed two different tasks (1) unsupervised - Dimension reduction and (2) supervised - Classification. Here, we compare standard and continuation training procedures. In particular, we are interested in the quality of the critical point to which our training methods converge. For this, we measure generalization performance using the test loss and accuracy. In all our experiments, we use the MNIST dataset (downsized to 6x6) and ADAM as solver for both standard and continuation approach. In Table-\ref{res:ae}, we show results when we embed homotopies for a three-layer autoencoder. We observe both NPC and PARC methods have better train and generalization performance. Similarly, in Table-\ref{res:clf}, we show results for a one-layer digit classifier, and the results are consistent, except for one data continuation task using PARC.

\begin{table}
    \centering
    \begin{tabular}{|p{1.5cm}|p{2.2cm}|c|c|}
    \hline
         Method & Homotopy & Train Loss & Test Loss  \\
         \hline
         Standard  &  ReLU & 0.0421 & 0.0422 \\
         (ADAM) &  Sigmoid & 0.0452 & 0.0458 \\
         \hline
         NPC &  h-ReLU & 0.042 & 0.042 \\
         & h-Sigmoid & 0.0401 & 0.0401 \\
         & h-Brightness & 0.0401 & 0.0402\\
         \hline
         PARC  &  h-ReLU & 0.040 & 0.040 \\
         (ours) &  h-Sigmoid & 0.0398 & 0.0399 \\
         &  h-Brightness & 0.0398 & 0.0398\\
    \hline
    \end{tabular}
    \caption{Three layer Autoencoder}
    \label{res:ae}
\end{table}

\begin{table}
    \centering
    \begin{tabular}{|p{1.2cm}|p{2.0cm}|c|c|p{2cm}|}
    \hline
         Method & Homotopy & Train Loss & Test Loss & Test Accuracy  \\
         \hline
         Standard (ADAM) &  ReLU & 0.64 & 0.675 & 0.78 \\
         \hline
         NPC &  h-ReLU &  0.58 & 0.59 & 0.814 \\
         &  h-Brightness & 0.51 & 0.53 & 0.827 \\
         \hline
             PARC  &  h-ReLU & 0.51 & 0.53 & 0.834 \\
         (ours) &  h-Brightness & 0.759 & 0.731 & 0.772 \\
    \hline
    \end{tabular}
    \caption{One layer classification network}
    \label{res:clf}
\end{table}

\section{Conclusion}
We proposed the Pseudo-arclength continuation  that introduces arclength parametrization \cite{pathak2024neural_phd} to the neural networks. Distinctly, we rethink the training of neural networks \cite{hershey2024rethinkingrelationshiprecurrentnonrecurrent, 10459970} as following a family of minima rather than standard solvers such as ADAM. We empirically observe better generalization performance for 4/5 optimization tasks. In the future, we hope to apply PARC to SOTA neural networks such as ResNet\cite{he2016deep_resnet, pathak2023sequentia12d}. We also want to derive some interpretations from the choice of $\lambda$ parameter and see how it affects the dynamics of training using bifurcation diagrams \cite{cont_method_Allgower}.
\bibliography{example_paper}
\bibliographystyle{icml2021}

\end{document}